\documentclass[journal=jcim,manuscript=article]{achemso}
\usepackage[utf8]{inputenc}
\usepackage{chemformula} 
\usepackage[T1]{fontenc} 
\usepackage{listings} 
\usepackage{url}
\usepackage{tikz}
\usepackage{color}
\usepackage{inconsolata}
\usepackage{float}

\definecolor{codebg}{HTML}{F7F7F7}
\definecolor{kw}{HTML}{204A87}     
\definecolor{str}{HTML}{4E9A06}    
\definecolor{cmt}{HTML}{8F5902}    
\definecolor{blt}{HTML}{5C35CC}    
\definecolor{num}{HTML}{7A7A7A}    
\definecolor{rule}{HTML}{CFCFCF}

\lstdefinestyle{skfp}{
  language=Python,
  backgroundcolor=\color{codebg},
  basicstyle=\small\ttfamily,
  keywordstyle=\color{kw}\bfseries,
  stringstyle=\color{str},
  commentstyle=\color{cmt}\itshape,
  emphstyle=\color{blt},
  numbers=left,
  numberstyle=\small\ttfamily\color{num},
  numbersep=8pt,
  frame=lines,
  rulecolor=\color{black},
  showstringspaces=false,
  breaklines=true,
  tabsize=4,
  columns=fullflexible,
  keepspaces=true,
  upquote=true,
  emph={self,True,False,None},
}
\author{Jakub Adamczyk}
\email{jadamczy@agh.edu.pl}
\altaffiliation{Contributed equally}

\author{Adam Staniszewski}
\altaffiliation{Contributed equally}

\affiliation{Faculty of Computer Science, AGH University of Krakow, 30-059 Cracow, Poland}

\title[Scikit-fingerprints: scikit-learn compatible chemoinformatics]
  {Scikit-fingerprints: Python library for scikit-learn compatible molecular fingerprints and chemoinformatics}

\abbreviations{}

\DeclareUnicodeCharacter{2009}{\,}
\begin{document}

\newcommand{\q}[1]{``#1''}
\renewcommand{\labelenumii}{\arabic{enumi}.\arabic{enumii}}

\begin{tocentry}
\begin{figure}[H]
  \centering
  \includegraphics[width=\textwidth]{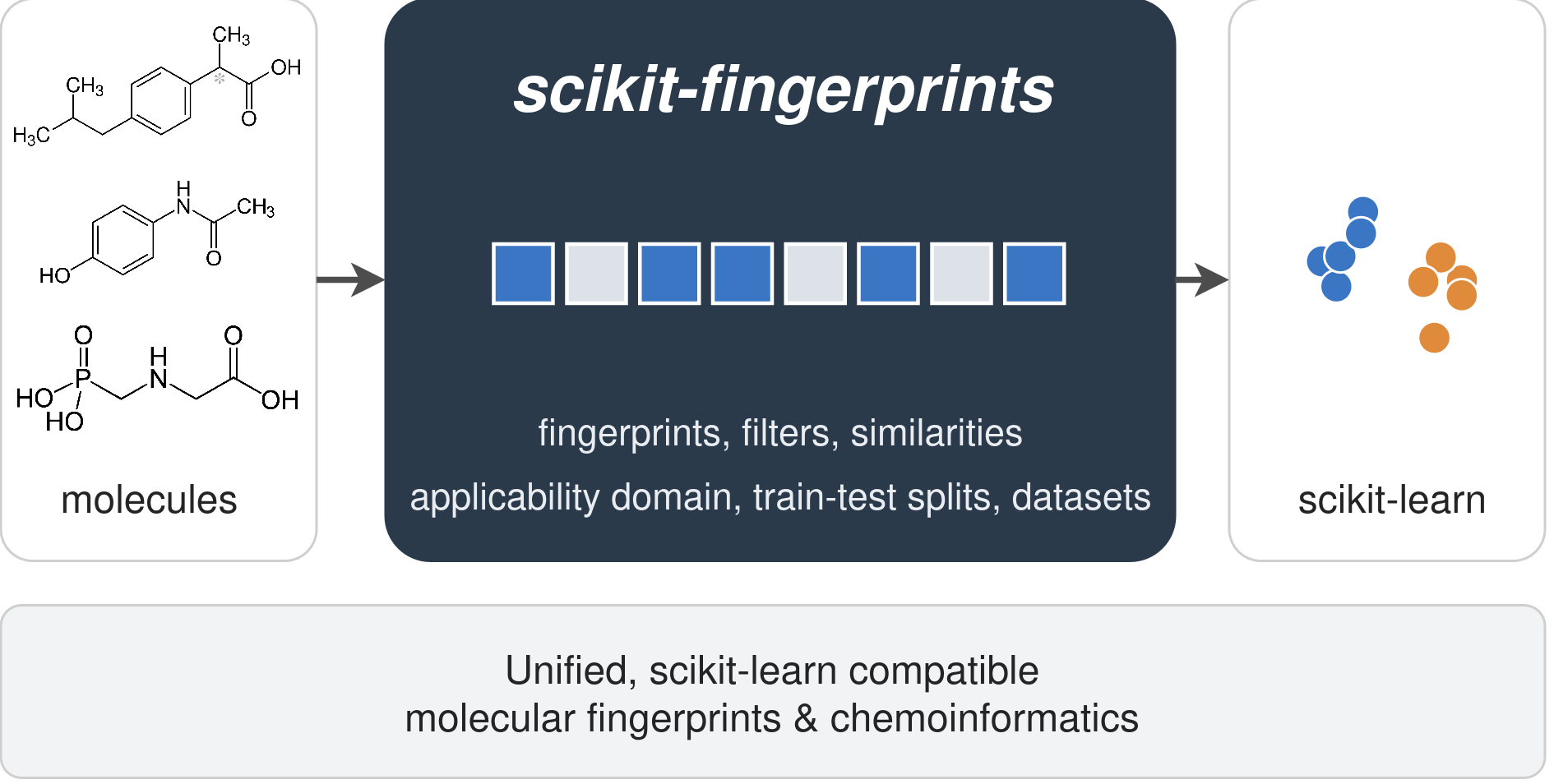}
\end{figure}
\end{tocentry}


\begin{abstract}
We present \textit{scikit-fingerprints}, a comprehensive, fully scikit-learn compatible library for molecular machine learning in Python, based on RDKit. Molecular fingerprints and related functionalities are workhorses of chemoinformatics, yet the widely used open-source frameworks are not compatible with the wider Python machine learning ecosystem based on scikit-learn conventions. \textit{scikit-fingerprints} closes this gap, bringing molecular fingerprints, molecular filters, similarity and distance measures, applicability domain estimation, data splitting strategies, and more under a single, familiar interface. Scikit-learn compatibility means that an entire chemoinformatics workflow, from a raw SMILES string to a deployable model, can be assembled from composable building blocks and can reuse the mature tooling of the surrounding ecosystem. The underlying RDKit code makes it familiar and extensible for custom chemoinformatics use cases. We put a strong focus on unified interfaces, ease of use, computational efficiency, customization, and extensibility. \textit{scikit-fingerprints} makes molecular machine learning faster to prototype, easier to reproduce, and simpler to deploy.
\end{abstract}

\section{Introduction and motivation}

Molecules are the fundamental objects of chemoinformatics. They are most commonly stored as molecular graphs, which must be turned into fixed-length numerical vectors before the majority of downstream algorithms, and machine learning (ML) models in particular, can process them. This vectorization is typically performed with molecular fingerprints, feature extraction algorithms that encode structural, topological, or physicochemical information about a molecule as a vector \cite{molecular_descriptors}. Fingerprints underpin a broad range of chemoinformatics tasks, including molecular property and activity prediction \cite{fingerprints_vs_gnn,fingerprints_comparison}, virtual screening and similarity searching \cite{virtual_screening_benchmark,similarity_search}, chemical space visualization \cite{similarity_maps,chemical_space_umap}, clustering and diversity picking \cite{butina,clustering}, and data splitting for reliable model evaluation \cite{moleculenet,ogb}. Fingerprint-based models remain highly competitive baselines, often matching or exceeding neural networks on molecular property prediction \cite{fingerprints_vs_gnn,moltop,fingerprints_peptides,apistox_ML,apistox_ML_CIKM}.

Building a QSAR/QSPR model, however, is rarely a single step. A realistic workflow reads and standardizes molecular structures, filters out undesirable compounds, computes one or several molecular fingerprints, splits the data to evaluate out-of-distribution generalization (e.g., via scaffold split), tunes hyperparameters, evaluates with domain-appropriate metrics, and finally assesses whether new predictions fall within the model's applicability domain \cite{molpipeline,applicability_domain}. Each of these steps has well-established methodology, yet in practice they are stitched together with ad hoc scripts, which is repetitive, error-prone, and hard to reproduce \cite{reproducibility}. As chemoinformatics workflows evolve from experimental scripts into production-grade deployments, there is a pressing need for proper software engineering practices and interoperability among the tools used.

In the Python ecosystem, scikit-learn \cite{scikit_learn} has become the de facto standard for tabular machine learning, valued for its consistent and widely adopted API \cite{sklearn_api}. A large family of compatible libraries follows the same conventions, including imbalanced-learn \cite{imbalanced_learn} for resampling, UMAP \cite{umap,umap2} for dimensionality reduction, and HDBSCAN \cite{hdbscan,hdbscan2} for clustering. This shared interface is precisely what makes complex pipelines composable and reusable. Chemoinformatics, unfortunately, has largely stood apart from it. The most widely used open-source toolkits - RDKit \cite{rdkit}, the Chemistry Development Kit (CDK) \cite{cdk}, and Open Babel \cite{open_babel} - are written in C++ or Java, expose object-oriented APIs tied to their own molecule types, and provide little to no support for the scikit-learn interface and for parallel computation. Their Python wrappers can be cumbersome to combine with ML tooling, resulting in brittle pipelines requiring significant custom code and duplication across QSAR/QSPR projects.

The first release of \textit{scikit-fingerprints} addressed one part of this gap by providing efficient, scikit-learn compatible molecular fingerprints \cite{scikit_fingerprints}. Since then, the library has grown well beyond fingerprinting into a general chemoinformatics toolkit, covering many functionalities applied either prior to computing molecular fingerprints, or based on those representations. In this work we present \textit{scikit-fingerprints} 2.0, whose scope now covers the whole molecular ML workflow inside the scikit-learn ecosystem. In addition to molecular fingerprints, the library provides molecular filters, similarity and distance measures, applicability domain estimators, scaffold-aware train-test splitters, efficient fingerprint-aware hyperparameter search, and other functionalities useful in building and evaluating QSAR/QSPR pipelines. It builds upon RDKit, with a focus on efficiency and parallelism, but also conforms to the scikit-learn API. This allows users to express experimental workflows as serializable pipelines, interoperable with the rest of the ecosystem and related libraries, and easily deployable into practical chemoinformatics applications.

\begin{sloppypar}
\textit{scikit-fingerprints} is fully open-source, distributed under the permissive MIT license, and available on PyPI. Code is hosted on GitHub at \url{https://github.com/MLCIL/scikit-fingerprints}. We provide extensive documentation and tutorials on GitHub and at \url{https://scikit-fingerprints.readthedocs.io/latest/}.
\end{sloppypar}

\section{Functionalities}

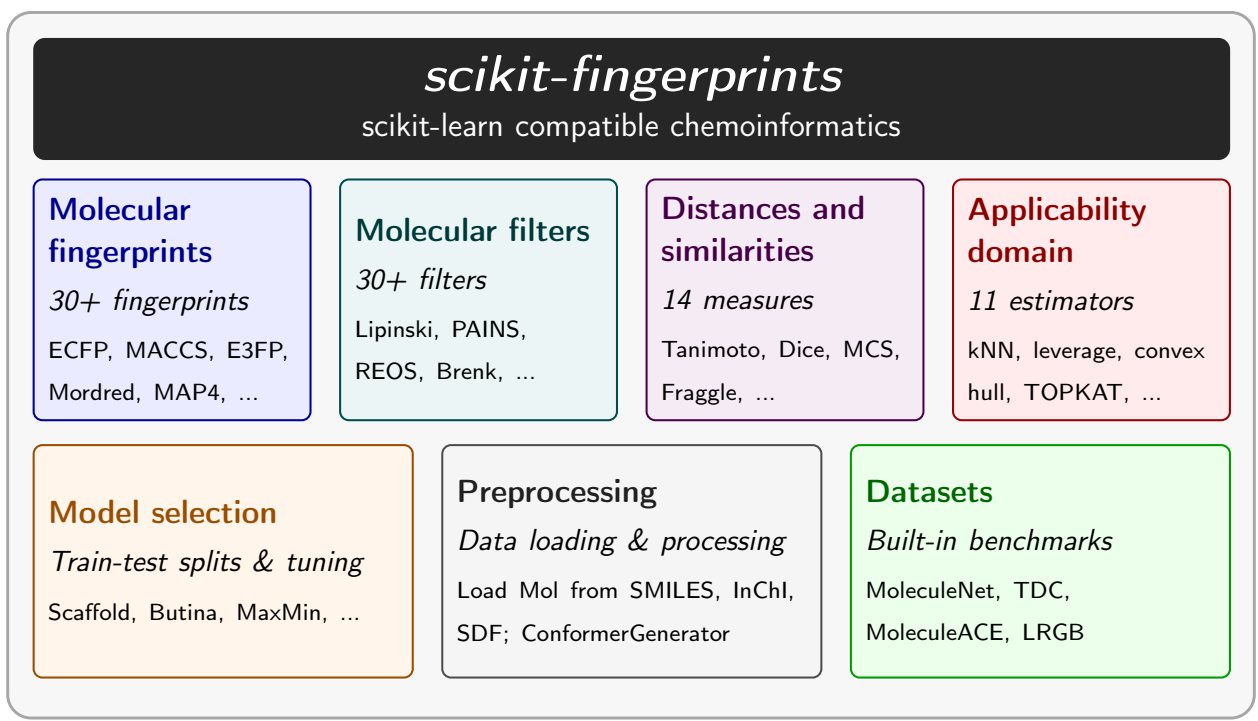
\begin{figure}[h!]
\centering
\resizebox{\linewidth}{!}{%
\begin{tikzpicture}[
  font=\sffamily,
  box/.style={anchor=north west, rounded corners=3pt, line width=0.7pt, inner sep=5pt, align=left},
  mod/.style n args={1}{box, draw=#1!60!black, fill=#1!8},
]
\draw[rounded corners=8pt, line width=1pt, draw=black!35, fill=black!3]
      (0,0) rectangle (15.0,-8.5);
 
\node[anchor=north west, rounded corners=4pt, draw=none, fill=black!85, text=white,
      align=center, inner sep=6pt, text width=13.98cm, minimum width=14.4cm,
      minimum height=0.95cm] (hdr) at (0.3,-0.3)
  {{\Large\bfseries\itshape scikit-fingerprints}\\[1pt]
   {\small scikit-learn compatible chemoinformatics}};
 
\node[mod={blue},   text width=2.98cm, minimum height=2.9cm] (fp)  at (0.30,-2.0)
  {{\small\bfseries\color{blue!55!black}Molecular \\fingerprints}\\[2pt]
   {\footnotesize\itshape 30+ fingerprints}\\[2pt]
   {\scriptsize ECFP, MACCS, E3FP, Mordred, MAP4, ...}};
\node[mod={teal},   text width=2.98cm, minimum height=2.9cm] (flt) at (3.9875,-2.0)
  {{\small\bfseries\color{teal!45!black}Molecular filters}\\[2pt]
   {\footnotesize\itshape 30+ filters}\\[2pt]
   {\scriptsize Lipinski, PAINS, REOS, Brenk, ...}};
\node[mod={violet}, text width=2.98cm, minimum height=2.9cm] (dist) at (7.675,-2.0)
  {{\small\bfseries\color{violet!55!black}Distances and similarities}\\[2pt]
   {\footnotesize\itshape 14 measures}\\[2pt]
   {\scriptsize Tanimoto, Dice, MCS, Fraggle, ...}};
\node[mod={red},    text width=2.98cm, minimum height=2.9cm] (ad) at (11.3625,-2.0)
  {{\small\bfseries\color{red!55!black}Applicability domain}\\[2pt]
   {\footnotesize\itshape 11 estimators}\\[2pt]
   {\scriptsize kNN, leverage, convex hull, TOPKAT, ...}};
 
\node[mod={orange}, text width=4.21cm, minimum height=2.8cm] (ms) at (0.30,-5.2)
  {{\small\bfseries\color{orange!60!black}Model selection}\\[2pt]
   {\footnotesize\itshape Train-test splits \& tuning}\\[2pt]
   {\scriptsize Scaffold, Butina, MaxMin, ...}};
\node[mod={gray},   text width=4.21cm, minimum height=2.8cm] (pre) at (5.2167,-5.2)
  {{\small\bfseries\color{gray!25!black}Preprocessing}\\[2pt]
   {\footnotesize\itshape Data loading \& processing}\\[2pt]
   {\scriptsize Load Mol from SMILES, InChI, SDF; ConformerGenerator}};
\node[mod={green},  text width=4.21cm, minimum height=2.8cm] (ds) at (10.1333,-5.2)
  {{\small\bfseries\color{green!40!black}Datasets}\\[2pt]
   {\footnotesize\itshape Built-in benchmarks}\\[2pt]
   {\scriptsize MoleculeNet, TDC, MoleculeACE, LRGB}};
\end{tikzpicture}%
}
\caption{Overview of \textit{scikit-fingerprints} functionalities.}
\label{fig:overview}
\end{figure}

All functionality lives in the \texttt{skfp} package, grouped into submodules that mirror the stages of a chemoinformatics workflow. See Figure~\ref{fig:overview} for an overview of functionalities.

Fingerprints and preprocessors are scikit-learn transformers, exposing the \texttt{.transform()} method. As they do not learn anything from the data, they are inherently stateless, so the \texttt{.fit()} method does not need to be called. All computations across different classes are parallelized with \texttt{joblib}, with controllable mini-batching and optional progress bars. Train-test splitters, similarity and distance measures, and quality metrics are plain functions, matching the scikit-learn signatures. Compatibility with the scikit-learn API is enforced by inheritance from scikit-learn base classes and verified by a suite of unit and integration tests, run on CI/CD with a matrix of Python versions and operating systems.

We note that \textit{scikit-fingerprints} is not merely an RDKit wrapper, but also implements a wide range of novel methods and functionalities, such as molecular filters based on physicochemical properties and applicability domain checkers. The following subsections describe each area, noting where components go beyond what RDKit provides directly. We also summarize the number of functionalities in different chemoinformatics libraries in Section ``Comparison with other software''.

\subsection{Molecular fingerprints}

Molecular fingerprints are feature extraction algorithms that turn a molecule into a fixed-length numerical vector that a machine learning model can consume. They encode structural, topological, or physicochemical information, embedding molecular graphs in a vector space, where tabular ML models can be applied \cite{molecular_descriptors}. The resulting vector representation itself is also commonly referred to as a fingerprint.

There is a large variety of fingerprint methods, since the choice of representations strongly influences downstream performance. This is particularly because the notion of molecular similarity in the resulting vector space depends on the algorithmic choices. The optimal choice for a given task cannot be known in advance, a problem further complicated by the fact that fingerprints themselves are configurable via various hyperparameters. For example, a popular Extended Connectivity FingerPrint (ECFP) \cite{ecfp} can optionally include chirality information, or count occurrences of subgraphs rather than just detect their presence (binary vs. count variant). As such, a practitioner typically wants to try many fingerprints, tune their hyperparameters, and also potentially build more complex pipelines, e.g., with feature selection or concatenating multiple fingerprints. \textit{scikit-fingerprints} is built around exactly this need, offering over 30 fingerprints behind a single, uniform interface.

Every fingerprint is a stateless scikit-learn transformer. All fingerprint classes inherit from \texttt{BaseFingerprintTransformer}, and accept a list of RDKit \texttt{Mol} objects or SMILES strings (parsed as molecules internally), returning either a dense NumPy array \cite{numpy} or a sparse SciPy CSR matrix \cite{scipy}. Behavioral options are set through constructor parameters, e.g., the output length for hashed fingerprints, or a binary vs. count variant. This uniformity is the key design decision: fingerprints can be swapped or combined easily within scikit-learn pipelines. Since the transformers are stateless, not learning any parameters or fitting training data, they carry an empty \texttt{.fit()} and can be dropped into pipelines, cross-validation, and grid search without special handling. See Listing~\ref{lst:fingerprints} for a code example.

The implemented fingerprints span all the major families used in chemoinformatics. Substructural fingerprints, such as MACCS \cite{maccs} and Klekota-Roth \cite{klekota_roth}, indicate the presence of predefined substructures, typically defined with SMARTS patterns, and yield directly interpretable, human-readable features. Hashed fingerprints enumerate local environments in a molecule and hash them into a vector of chosen length. These include, for example, ECFP \cite{ecfp} (also known as the Morgan fingerprint), Atom Pair \cite{atom_pair}, and Topological Torsion \cite{topological_torsion}. Descriptor-based fingerprints collect scalar physicochemical or graph-theoretic quantities into a vector, ranging from electrotopological EState indices \cite{estate} to the comprehensive Mordred descriptor set \cite{mordred}. A substantial subset of the fingerprints is conformer-based, capturing spatial geometry rather than the molecular graph alone. These include, for example, GETAWAY \cite{getaway}, E3FP \cite{e3fp}, and MAP4 \cite{map4}. This coverage includes all fingerprints commonly available in established toolkits alongside further additions, giving users a single place to obtain and compare them.

Computing fingerprints over large molecular datasets is often the most expensive step of a workflow, so efficiency is treated as a first-class concern. Because molecules are processed independently, fingerprint computation is embarrassingly parallel. Every transformer exposes an \texttt{n\_jobs} parameter and parallelizes across CPU cores using Joblib \cite{joblib}, which also supports distributed computation on HPC clusters via a Dask backend \cite{dask}. The underlying structural matching relies on RDKit's optimized C++ routines \cite{rdkit}. Since fingerprints are typically long and extremely sparse \cite{scikit_fingerprints}, particularly substructural and hashed ones, sparse output sharply reduces memory use, which is especially valuable during hyperparameter search when several copies of the feature matrix may coexist.

Following scikit-learn convention, we assume that data fits in memory, which for sparse fingerprints allows processing at least hundreds of thousands of molecules comfortably on commodity hardware. We note that memory is used primarily by RDKit \texttt{Mol} objects, rather than fingerprint arrays. For larger-than-memory data, we recommend using swap space on SSD drives, or processing molecules manually in batches.

Interpretability is supported wherever it is meaningful. Fingerprints with human-readable features implement the standard \texttt{.get\_feature\_names\_out()} method, so their outputs connect directly to scikit-learn feature importance measures. As this is a standard API, it also directly integrates with external libraries and explanation tools such as SHAP \cite{shap}, allowing model behavior to be traced back to concrete substructures or descriptors.

Some fingerprints, like Mordred \cite{mordred}, may produce missing values (NaN) for certain molecules, such as salts with disconnected components. Following scikit-learn convention, we do not handle those internally, instead returning them as-is. This gives the user maximal flexibility, allowing the use of any imputation strategy or removing such features altogether.

\begin{lstlisting}[caption={Molecular fingerprints example.},label={lst:fingerprints},float=h!]
from skfp.fingerprints import ECFPFingerprint
 
smiles_list = ["O=C(O)c1ccccc1O", "CCO", "[C-]#N"]
fp = ECFPFingerprint(count=True, n_jobs=-1)
X = fp.transform(smiles_list)  # SMILES in, NumPy array out
\end{lstlisting}

\subsection{Molecular filters}

Molecular filters remove compounds that are undesirable for a given task, such as retaining only drug-like compounds or removing known reactive groups. This is a standard step in many chemoinformatics workflows, particularly virtual screening and compound library design. Filters can be roughly divided into two groups: those based on allowed ranges of physicochemical properties, and those detecting disallowed substructures. The former encode primarily drug-likeness or lead-likeness rules based on simple molecular properties, including Lipinski's Rule of Five \cite{lipinski}, Ghose \cite{ghose}, Veber \cite{veber}, and REOS \cite{reos}, among others. We note that no such filters are implemented directly in RDKit, whereas \textit{scikit-fingerprints} implements over a dozen. Substructural filters flag problematic chemical motifs, most notably PAINS \cite{pains} for pan-assay interference compounds, with others including, e.g., Brenk \cite{brenk} and Glaxo \cite{glaxo} rule sets, defined with SMARTS patterns.

\textit{scikit-fingerprints} provides over 30 filters as transformers that accept molecules and return the subset passing the rules. See Listing~\ref{lst:filters} for a code example. As filtering is often an interactive process, detailed results on passing or failing particular filter conditions can also be returned. Since filters are often designed for specific applications, such as pesticides \cite{hao_pesticides_filter,apistox} or RNA-targeting small compounds \cite{rna_binding}, implementing custom ones is particularly easy. It requires only inheriting from the base filter class and overriding the \texttt{\_apply\_mol\_filter()} method.

Filtering changes the number of samples, which the standard scikit-learn \texttt{.transform()} interface does not account for, as it returns only the features. To keep labels aligned with the retained molecules, every filter additionally provides the \texttt{.transform\_x\_y()} method, as inspired by the feature-engine library \cite{feature_engine}. It accepts molecules and their labels, and returns both filtered consistently. This avoids manual index bookkeeping, which is a common source of subtle errors in supervised workflows, and is particularly useful when several filters are chained in sequence.

\begin{lstlisting}[caption={Molecular filters example.},label={lst:filters},float=h!]
from skfp.filters import LipinskiFilter

smiles_list = ["O=C(O)c1ccccc1O", "CCO", "[C-]#N"]
filt = LipinskiFilter()
smiles_passing = filt.transform(smiles_list)  # subset that passes
\end{lstlisting}

\subsection{Distances and similarities}

Similarity and distance measures are central to virtual screening, clustering, and diversity analysis. RDKit provides such measures, but only for its own fingerprint objects, which prevents their direct use with the distance-based algorithms of the scikit-learn ecosystem. \textit{scikit-fingerprints} implements them as functions, primarily based on NumPy and SciPy arrays, making them fully interoperable with scikit-learn. These include classic similarities like Tanimoto, Dice, Sokal-Sneath, or Kulczynski, as well as measures based on direct comparison of molecular graphs, such as Maximum Common Subgraph (MCS) \cite{mcs} and Fraggle \cite{fraggle} similarities, with over a dozen measures in total. Each one is available in similarity and distance forms and, where applicable, in binary and count variants.

Because the distance functions have standard signatures, they can be passed as custom metrics to, for example, scikit-learn k-nearest neighbors, UMAP, HDBSCAN, and other methods based on nearest neighbors. This aligns the behavior of those algorithms with fingerprint representations throughout the ecosystem.

Another application is the computation of pairwise similarity matrices, which can be used as kernel matrices \cite{graph_kernels} for SVMs and other kernel methods, as well as for diversity analysis, clustering, and other downstream tasks. We implemented a highly optimized version of such bulk computation of $N \times N$ matrices with sparse matrix operations, utilizing the fact that molecular fingerprints are highly sparse, with often just a few percent of non-zero entries \cite{scikit_fingerprints}. This results in significant speedups over naive loops, due to vectorization and better CPU utilization, allowing scaling to larger datasets. See Listing~\ref{lst:distances} for a code example.

\begin{lstlisting}[caption={Bulk similarity computation example.},label={lst:distances},float=h!]
from skfp.distances import bulk_tanimoto_binary_similarity
from skfp.fingerprints import ECFPFingerprint

smiles_list = ["O=C(O)c1ccccc1O", "CCO", "[C-]#N"]
fp = ECFPFingerprint()
X = fp.transform(smiles_list)
sims = bulk_tanimoto_binary_similarity(X, X) # NxN matrix
\end{lstlisting}

\subsection{Model selection}

Reliable evaluation requires splits that reflect the distribution shift a model will encounter in practice. \textit{scikit-fingerprints} provides five scaffold-aware and diversity-aware splitting strategies as functions matching scikit-learn's \texttt{train\_test\_split} signature: deterministic and randomized Bemis-Murcko scaffold splits \cite{bemis_murcko,ogb}, Butina clustering split \cite{butina}, MaxMin diversity split \cite{maxmin}, and a PubChem-based time split \cite{pubchem}. Each is available in both train-test and train-validation-test forms. See Listing~\ref{lst:splitting} for a code example.

\begin{lstlisting}[caption={Scaffold train-test split example.},label={lst:splitting},float=h!]
from skfp.model_selection import scaffold_train_test_split

smiles = ["c1ccccc1", "C1CCCCC1", "CCO", "CCN", "CCCl", "CCBr"]

train_smiles, test_smiles = scaffold_train_test_split(smiles, test_size=2)
\end{lstlisting}

\begin{sloppypar}
Hyperparameter tuning is a common bottleneck for QSAR/QSPR pipelines, because scikit-learn assumes that a \texttt{.transform()} call is cheap and can be recomputed for every combination of hyperparameters. When tuning hyperparameters of molecular fingerprints and a downstream model, e.g., LightGBM \cite{lightgbm}, the naive combination of \texttt{Pipeline} and \texttt{GridSearchCV} recomputes the fingerprint for every estimator configuration. \textit{scikit-fingerprints} addresses this with a nested search loop. Two classes implement it: \texttt{FingerprintEstimatorGridSearch} for exhaustive search, and \texttt{FingerprintEstimatorRandomizedSearch} for randomized search. In both, the fingerprint is computed once per fingerprint configuration and reused across all estimator configurations; when only estimator hyperparameters are tuned, it is computed a single time. The savings grow with the cost of the fingerprint and the size of the estimator grid, as we show in the Benchmarks section.
\end{sloppypar}

QSAR models are frequently used for virtual screening, where the top-k ranking is of interest, rather than the overall classification performance. \textit{scikit-fingerprints} implements dedicated virtual screening measures, including enrichment factor, RIE \cite{rie}, and BEDROC \cite{bedroc}. Thanks to scikit-learn compatibility, they can, for example, be used directly as optimization targets in hyperparameter tuning.

\subsection{Preprocessing utilities}

Preprocessing transformers cover the entry and exit points of a pipeline. Input-output classes convert between molecules and SMILES, InChI, SDF, and amino-acid sequences, for example, \texttt{MolFromSmilesTransformer} and \texttt{MolToInchiTransformer}. Since a molecule can have many valid SMILES forms and real datasets contain questionable structures, \texttt{MolStandardizer} applies the widely recommended RDKit standardization steps to improve data quality at the start of a workflow. See Listing~\ref{lst:loading} for an example.

Conformer-based fingerprints require 3D structures, which can be generated with the \texttt{ConformerGenerator} class using the \texttt{.transform()} method. It wraps the ETKDGv3 algorithm \cite{etkdg} with default settings that balance speed on easy molecules against robustness on difficult ones. Conformers for 3D fingerprints can also be provided with external tools, but our implementation is designed for easy, robust usage and is also fully parallelized.

Real-world molecular datasets frequently contain entries that RDKit cannot parse, such as malformed SMILES or hypervalent atoms that fail default sanitization. By default, such inputs raise an error, which makes problems explicit and prevents corrupting a dataset. As such cases are usually rare, preprocessing classes accept a \texttt{valid\_only} parameter, which when set to \texttt{True} silently omits problematic molecules from the output. Because this changes the number of samples, those classes also implement the \texttt{.transform\_x\_y()} method, keeping labels aligned with the retained molecules. This makes it easy to build robust ingestion pipelines over noisy data sources, while retaining the strict behavior as the default.

\begin{lstlisting}[caption={Loading molecules from SMILES strings.},label={lst:loading},float=h!]
from skfp.preprocessing import MolFromSmilesTransformer

smiles = ["c1ccccc1", "C1CCCCC1", "CCO", "CCN", "CCCl", "CCBr"]

mol_from_smiles = MolFromSmilesTransformer(valid_only=True)
mols = mol_from_smiles.transform(smiles)
\end{lstlisting}

\subsection{Applicability domain checks}

A model's predictions are trustworthy only for molecules resembling its training data. Applicability domain (AD) analysis quantifies this and is essential for responsible deployment, yet it has no native RDKit support. \textit{scikit-fingerprints} implements 11 AD estimators from scratch, including distance and density approaches such as k-nearest neighbors and distance-to-centroid, geometric approaches such as bounding box, PCA bounding box, and convex hull, and statistical approaches such as leverage, Hotelling's $T^2$ test, and the TOPKAT descriptor-range method \cite{applicability_domain}.

As those are dataset-dependent estimators, they implement the scikit-learn estimator interface with \texttt{.fit()} and \texttt{.predict()}. Fitted on training data, the estimator determines whether new molecules fall inside the domain, a check useful for flagging potentially unreliable predictions in production. Further, many methods provide a per-sample measure of prediction reliability, e.g., the leverage value \cite{leverage}, and they can be computed using the \texttt{.score\_samples()} method.

\subsection{Built-in datasets}

Fair benchmarking depends on standardized datasets and splits. \textit{scikit-fingerprints} hosts curated datasets on the HuggingFace Hub \cite{huggingface} with automatic downloading, caching, and Parquet-based compression, loaded through scikit-learn-style functions such as \texttt{load\_bbbp()}.

The current version integrates the widely used MoleculeNet benchmark \cite{moleculenet}, the Therapeutics Data Commons (TDC) \cite{tdc}, the Long Range Graph Benchmark (LRGB) peptide datasets \cite{lrgb}, and MoleculeACE for activity-cliff evaluation \cite{moleculeace}. Data can be returned as SMILES and labels or as a Pandas DataFrame \cite{pandas}. Predefined train-test splits are also provided where possible for standardized benchmarking. See Listing~\ref{lst:datasets} for a code example. 

\begin{lstlisting}[caption={Example of loading a built-in MoleculeACE ChEMBL 204 Ki dataset.},label={lst:datasets},float=h!]
from skfp.datasets.moleculeace import load_chembl204_ki

# downloads from HuggingFace Hub as necessary
smiles_list, labels = load_chembl204_ki()
\end{lstlisting}

\section{Comparison with other software}
\label{section_comparison_other_software}

There are several alternative open source packages that perform similar functions to scikit-fingerprints. We summarize the functionalities offered in Table~\ref{tab_other_software}. Overall, scikit-fingerprints offers the widest range of capabilities, combining the largest collection of molecular fingerprints with molecular filters, applicability domain checkers, and integrated benchmark datasets, which are not jointly available elsewhere.

CDK \cite{cdk} and Open Babel \cite{open_babel} are mature and feature-rich toolkits, but they are not compatible with scikit-learn. Further, CDK is Java-based and unavailable on PyPI, while Open Babel's copyleft GPL-2.0 license restricts use in proprietary settings. RDKit \cite{rdkit} is the most comprehensive general-purpose toolkit and serves as the computational backend of scikit-fingerprints, but its object-oriented C++ API is not scikit-learn compatible, and it provides neither applicability domain analysis nor curated datasets. Among the Python-native, scikit-learn compatible libraries, scikit-mol \cite{scikit_mol} and MolPipeline \cite{molpipeline} share our design philosophy, but address narrower parts of the workflow, with considerably fewer fingerprints and no filters or integrated datasets. scikit-fingerprints is therefore the first package to cover the full QSAR/QSPR workflow under a single permissively licensed and fully scikit-learn compatible interface.

\begin{table}[]
\caption{Comparison of capabilities of different chemoinformatics software.}
\resizebox{\textwidth}{!}{
\begin{tabular}{|c|c|c|c|c|c|c|}
\hline
\textbf{} & \textbf{CDK} & \textbf{Open Babel} & \textbf{RDKit} & \textbf{scikit-mol} & \textbf{MolPipeline} & \textbf{scikit-fingerprints} \\ \hline
Language & Java & C++ & C++ & Python & Python & Python \\ \hline
pip-installable & No & Yes & Yes & Yes & Yes & Yes \\ \hline
Last PyPI update & Never & 2026 & 2026 & 2025 & 2026 & 2026 \\ \hline
Fingerprints & 13 & 7 & 25 & 8 & 6 & 34 \\ \hline
Molecular filters & 0 & 0 & 15 & 0 & 0 & 31 \\ \hline
\begin{tabular}[c]{@{}c@{}}Applicability\\ domain checkers\end{tabular} & 0 & 0 & 0 & 11 & 0 & 11 \\ \hline
\begin{tabular}[c]{@{}c@{}}Integrated\\ datasets\end{tabular} & No & No & No & No & No & Yes \\ \hline
\begin{tabular}[c]{@{}c@{}}scikit-learn\\ compatible\end{tabular} & No & No & No & Yes & Yes & Yes \\ \hline
Parallelism & No & Yes & Partial & Yes & Yes & Yes \\ \hline
\begin{tabular}[c]{@{}c@{}}Easy proprietary\\ usage\end{tabular} & \begin{tabular}[c]{@{}c@{}}Yes\\ (LGPL-2.1)\end{tabular} & \begin{tabular}[c]{@{}c@{}}No\\ (GPL-2.0)\end{tabular} & \begin{tabular}[c]{@{}c@{}}Yes\\ (BSD-3-Clause)\end{tabular} & \begin{tabular}[c]{@{}c@{}}Yes\\ (LGPL-3.0)\end{tabular} & \begin{tabular}[c]{@{}c@{}}Yes\\ (MIT)\end{tabular} & \begin{tabular}[c]{@{}c@{}}Yes\\ (MIT)\end{tabular} \\ \hline
\end{tabular}
}
\label{tab_other_software}
\end{table}

\section{Usage examples}

Here, we present larger usage examples, as the value of unified interfaces is greatest in end-to-end pipelines. In particular, full chemoinformatics workflows can be easily implemented with \textit{scikit-fingerprints}, covering various tasks such as QSAR/QSPR, virtual screening, visualization, and more.

In Listing~\ref{lst:herg}, we present a complete classification pipeline for the hERG dataset by Karim et al. \cite{herg_karim} from the Therapeutics Data Commons (TDC) benchmark \cite{tdc}. This dataset is a binary classification task, with molecules labeled as hERG ($<10\,\mu\text{M}$) and non-hERG ($\geq10\,\mu\text{M}$) blockers. We load the data (automatically downloading it from HuggingFace Hub), convert SMILES to molecules, perform a scaffold train-test split, compute count ECFP fingerprints, train a Random Forest classifier, and evaluate using the AUROC metric. The entire workflow integrates directly with scikit-learn as a \texttt{Pipeline} object, which is finally retrained on the whole dataset and serialized to disk with Joblib. It could then be deployed to a production QSAR application to serve predictions via a web server, for example.

\begin{lstlisting}[caption={End-to-end QSAR hERG classification pipeline.},label={lst:herg},float=h!]
import joblib
from sklearn.ensemble import RandomForestClassifier
from sklearn.metrics import roc_auc_score
from sklearn.pipeline import make_pipeline

from skfp.datasets.tdc import load_herg_karim
from skfp.fingerprints import ECFPFingerprint
from skfp.model_selection import scaffold_train_test_split
from skfp.preprocessing import MolFromSmilesTransformer

smiles, y = load_herg_karim()
smiles_train, smiles_test, y_train, y_test = scaffold_train_test_split(
    smiles, y, test_size=0.2
)

pipeline = make_pipeline(
    MolFromSmilesTransformer(),
    ECFPFingerprint(count=True),
    RandomForestClassifier(n_jobs=-1, random_state=0),
)
pipeline.fit(smiles_train, y_train)

y_pred = pipeline.predict_proba(smiles_test)[:, 1]
auroc = roc_auc_score(y_test, y_pred)
print(f"AUROC: {auroc:.2%}")

pipeline.fit(smiles, y)
joblib.dump(pipeline, "herg_model.joblib")
\end{lstlisting}

As a more involved pipeline, in Listing~\ref{lst:conformer-qsar} we present a conformer-based QSAR pipeline for regression, which also integrates with other libraries from the scikit-learn ecosystem. We first load a generic CSV file with ``SMILES'' and ``target'' columns. Then we generate conformations with the ETKDGv3 algorithm, perform a Butina train-test split, compute and concatenate E3FP fingerprints and RDKit 2D descriptors, remove highly correlated features with feature-engine \cite{feature_engine}, train a LightGBM regressor \cite{lightgbm}, and evaluate using the MAE metric.

\begin{lstlisting}[caption={Conformer-based QSAR regression.},label={lst:conformer-qsar},float=h!]
import pandas as pd
from feature_engine.selection import DropCorrelatedFeatures
from lightgbm import LGBMRegressor
from sklearn.metrics import mean_absolute_error
from sklearn.pipeline import make_pipeline, make_union

from skfp.fingerprints import E3FPFingerprint, RDKit2DDescriptorsFingerprint
from skfp.model_selection import butina_train_test_split
from skfp.preprocessing import ConformerGenerator, MolFromSmilesTransformer

df = pd.read_csv("dataset.csv")  # columns: "SMILES", "target"
smiles = df["SMILES"]
labels = df["target"]

mol_from_smiles = MolFromSmilesTransformer(n_jobs=-1)
conf_gen = ConformerGenerator(optimize_force_field="MMFF94", n_jobs=-1)

mols = mol_from_smiles.transform(smiles)
mols = conf_gen.transform(mols)

mols_train, mols_test, y_train, y_test = butina_train_test_split(
    mols, labels, test_size=0.2
)

pipeline = make_pipeline(
    make_union(
        E3FPFingerprint(n_jobs=-1),
        RDKit2DDescriptorsFingerprint(n_jobs=-1),
    ),
    DropCorrelatedFeatures(threshold=0.9),
    LGBMRegressor(n_jobs=-1, random_state=0),
)
pipeline.fit(mols_train, y_train)

y_pred = pipeline.predict(mols_test)
mae = mean_absolute_error(y_test, y_pred)
print(f"MAE: {mae:.3f}")
\end{lstlisting}

Ligand-based virtual screening pipelines are commonly applied to filter down large libraries. \textit{scikit-fingerprints} offers parallelization with mini-batching, molecular filters, sparse fingerprints for memory usage reduction, and other functionalities useful for this purpose. In Listing~\ref{lst:screening}, we present such a pipeline for a generic bioactivity screening, modeled as binary classification. We apply REOS and PAINS filters \cite{reos,pains}, perform a MaxMin split \cite{maxmin}, calculate sparse PubChem fingerprints \cite{pubchem_fp}, train a logistic regression classifier using a fast solver for sparse matrices, and finally evaluate using the BEDROC metric \cite{bedroc}, which checks early enrichment capabilities. This pipeline is applied directly to SMILES, creating RDKit \texttt{Mol} objects internally as mini-batches. This minimizes memory usage, as, at scale, those objects take up significantly more memory than SMILES strings.

\begin{lstlisting}[caption={Ligand-based virtual screening pipeline.},label={lst:screening},float=h!]
import pandas as pd
from sklearn.linear_model import LogisticRegression
from sklearn.pipeline import make_pipeline

from skfp.filters import PAINSFilter, REOSFilter
from skfp.fingerprints import PubChemFingerprint
from skfp.metrics import bedroc_score
from skfp.model_selection import maxmin_train_test_split

df = pd.read_csv("dataset.csv")
smiles = df["SMILES"]
y = df["label"]

# verbose option shows progress bars
filt_reos = REOSFilter(n_jobs=-1, batch_size=1000, verbose=True)
filt_pains = PAINSFilter(n_jobs=-1, batch_size=1000, verbose=True)

smiles_filtered, y_filtered = filt_reos.transform_x_y(smiles, y)
smiles_filtered, y_filtered = filt_pains.transform_x_y(smiles_filtered, y_filtered)

smiles_train, smiles_test, y_train, y_test = maxmin_train_test_split(
    smiles_filtered, y_filtered, test_size=0.2
)

pipeline = make_pipeline(
    PubChemFingerprint(sparse=True, batch_size=1000, n_jobs=-1),
    LogisticRegression(solver="newton-cg"),  # solver for sparse problems
)
pipeline.fit(smiles_train, y_train)

y_pred = pipeline.predict_proba(smiles_test)[:, 1]
bedroc = bedroc_score(y_test, y_pred)
print(f"BEDROC: {bedroc:.3f}")
\end{lstlisting}

\clearpage

Lastly, in Listing~\ref{lst:umap} we show a data exploration scenario, where the goal is to visualize a molecular dataset on a 2D plane using UMAP \cite{umap,umap2}. \textit{scikit-fingerprints} offers distances and similarities based on NumPy arrays, with interfaces integrating with the rest of the scikit-learn ecosystem. We load the BACE dataset from the MoleculeNet benchmark \cite{moleculenet}, compute count ECFP fingerprints, and then pass the Tanimoto distance for count vectors \cite{tanimoto} to UMAP for dimensionality reduction.

\begin{lstlisting}[caption={Molecular dataset visualization.},label={lst:umap},float=h!]
import matplotlib.pyplot as plt
import umap

from skfp.datasets.moleculenet import load_bace
from skfp.distances import tanimoto_count_distance
from skfp.fingerprints import ECFPFingerprint

smiles, y = load_bace()
fp = ECFPFingerprint(count=True)
X = fp.transform(smiles)

embedding = umap.UMAP(metric=tanimoto_count_distance).fit_transform(X)

plt.scatter(embedding[:, 0], embedding[:, 1], c=y)
plt.show()
\end{lstlisting}

\section{Benchmarks}

We evaluate \textit{scikit-fingerprints} in a few selected use cases. All experiments in this section are based on the representative HIV dataset of the MoleculeNet benchmark \cite{moleculenet}. It was chosen since it is large enough (over $41,000$ compounds) to evaluate scalability on typical QSAR/QSPR workflows. Experiments were performed on a PC with an Intel Core i9-13900H CPU (14 cores, 20 threads) and 64 GB RAM. All reported measurements are averages of 5 runs.

\subsection{Hyperparameter tuning}

\begin{sloppypar}
As described in the Model selection section, \textit{scikit-fingerprints} implements the \texttt{FingerprintEstimatorGridSearch} class. Whereas the typical combination of \texttt{GridSearchCV} and \texttt{Pipeline} in scikit-learn recomputes fingerprints for every parameter combination and cross-validation fold, our implementation optimizes this, reusing fingerprints with given hyperparameters when tuning downstream classifiers. Gains are particularly strong when tuning fingerprints that are more expensive to compute, and with larger grids of downstream models.
\end{sloppypar}

We use a combination of the RDKit fingerprint \cite{rdkit} and a Random Forest (RF) classifier, with two hyperparameter grids for tuning:
\begin{itemize}
    \item small: fingerprint \texttt{\{ ``count'': [False, True], ``fp\_size'': [1024, 2048] \}}, Random Forest \texttt{\{ ``min\_samples\_split'': [2, 5, 10] \}}

    \item large: fingerprint \texttt{\{ ``count'': [False, True], ``fp\_size'': [512, 1024, 2048] \}}, Random Forest \texttt{\{ ``min\_samples\_split'': [2, 5, 10], ``max\_depth'': [None, 10, 20], ``n\_estimators'': [100, 200, 500] \}}
\end{itemize}

For the small grid, this results in 4 combinations for the fingerprint and 3 for RF, so with 5-fold CV the scikit-learn implementation computes the fingerprint 60 times, and \textit{scikit-fingerprints} only 4 times. For the large grid, there are 6 and 27 combinations, respectively, which gives 810 fingerprint computations for scikit-learn, and 6 for \textit{scikit-fingerprints}.

\begin{figure}[H]
  \centering
  \includegraphics[width=0.8\linewidth]{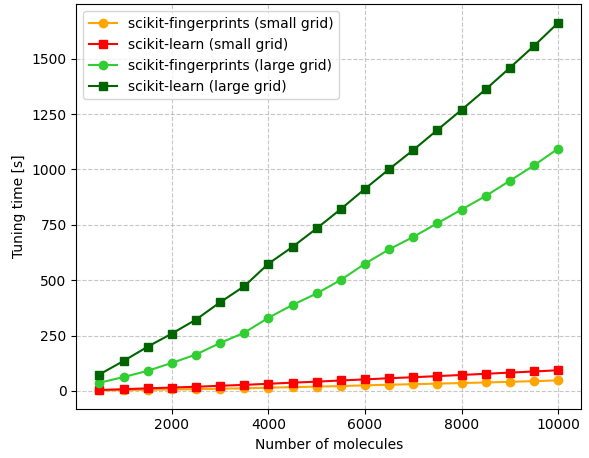}
  \caption{Hyperparameter tuning time.}
  \label{fig:hp-tuning-benchmark}
\end{figure}

Figure~\ref{fig:hp-tuning-benchmark} shows total tuning time as a function of dataset size. Our implementation is faster in every case, but the gains are stronger for the large grid, as expected. For 10,000 molecules, \textit{scikit-fingerprints} is faster by almost 9.5 minutes.

\subsection{PubChem fingerprint calculation}

The PubChem (CACTVS) fingerprint \cite{pubchem_fp} is an 881-bit substructure-key fingerprint, designed for similarity searching, and exposed through the PUG REST web API. However, this implementation has considerable downsides: it works only for compounds in PubChem, requires first fetching the CID number based on SMILES, requires a large number of HTTP requests, is frequently overloaded and returns errors requiring retries, and allows only limited parallelization per client due to rate limiting. Together, these result in very high computation time to embed whole datasets.

\begin{figure}[H]
  \centering
  \includegraphics[width=0.7\linewidth]{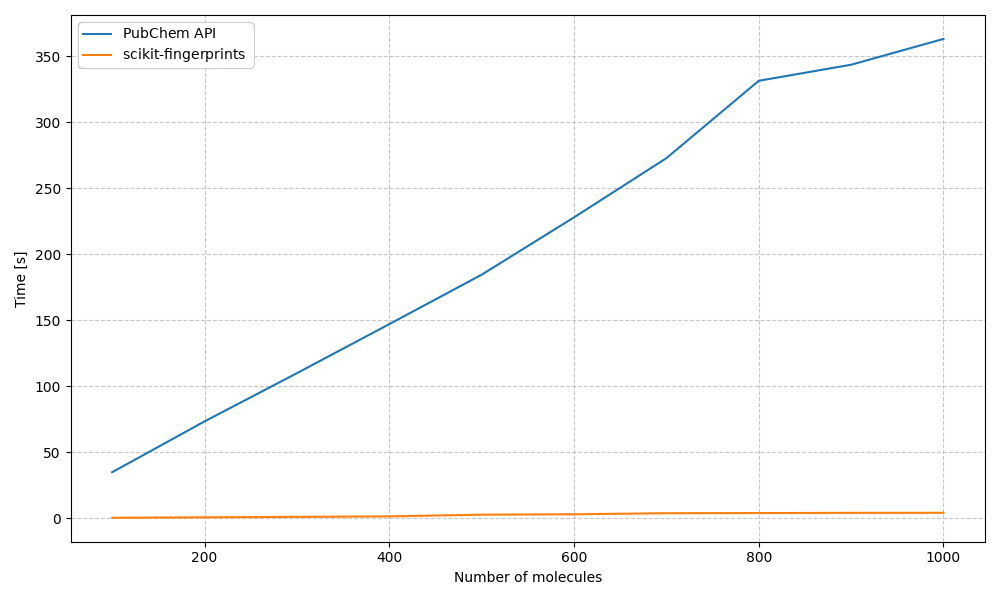}
  \caption{PubChem fingerprint computation.}
  \label{fig:pubchem_fp}
\end{figure}

As key definitions are publicly available, we implemented them in the \texttt{PubChemFingerprint} class, using SMARTS patterns. Our implementation is also fully parallelized, and we also offer a count-based variant, as many binary keys rely on counting atoms, e.g., the number of carbons $C \geq 2$, $ C \geq 4$, $C \geq 8 $, $C \geq 16$. Our local implementation is considerably faster, primarily thanks to no network latency, as shown in Figure~\ref{fig:pubchem_fp}, allowing scaling to much larger datasets. We measure only the time to calculate fingerprints based on CID numbers, so in practice the total API time would be roughly twice as long.

\subsection{Bulk Tanimoto similarity computation}

Computing bulk Tanimoto similarities between molecules, resulting in an $N \times K$ similarity matrix, arises in many places in chemoinformatics. Examples include evaluating similarity of known active ligands to compounds in a screening catalog, or visualizing the distribution of similarities between molecules in a dataset (resulting in an $N \times N$ matrix) or between datasets. As molecular fingerprints are very sparse, and most similarity measures for binary vectors use only bit-wise AND / OR / SUM operations, such bulk computation can be implemented using sparse CSR matrix operations. Such an approach is not available in RDKit, which only supports one-to-many similarity evaluations, resulting in function call overhead.

\begin{figure}[H]
  \centering
  \includegraphics[width=\linewidth]{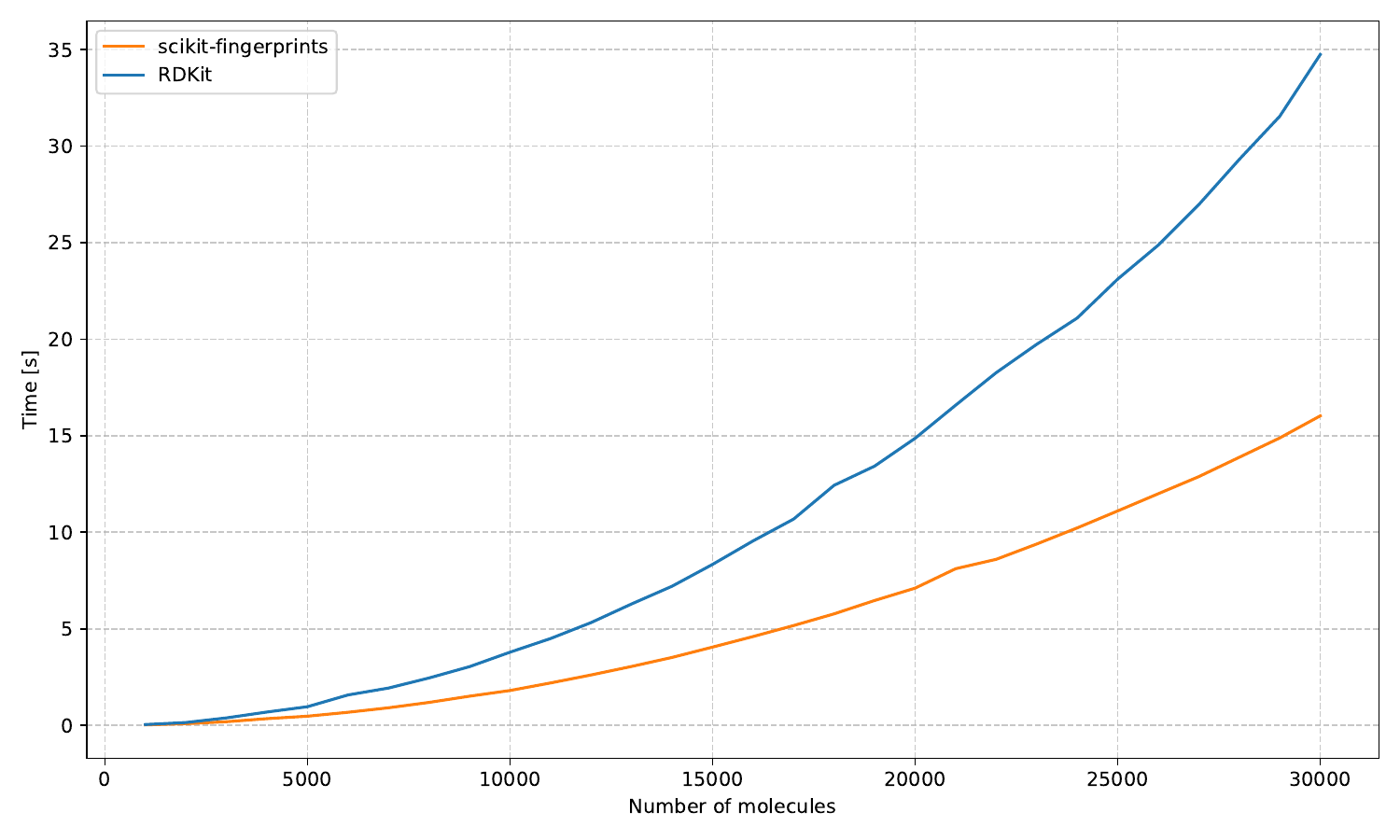}
  \caption{Bulk Tanimoto similarity computation.}
  \label{fig:bulk_tanimoto}
\end{figure}

In Figure~\ref{fig:bulk_tanimoto}, we compare the speed of \textit{scikit-fingerprints} and RDKit in bulk $N \times N$ similarity matrix computation. Our implementation is considerably faster for this type of task, with favorable scalability particularly for larger data sizes.

\subsection{Large-scale fingerprint computation}

Molecular fingerprints are nowadays used for exploration and virtual screening of large-scale chemical spaces. For such use cases, parallelization, mini-batching, and sparse representations in \textit{scikit-fingerprints} are directly useful. As an example, we calculate 2048-bit ECFP4 fingerprints on COCONUT \cite{coconut}, ChEMBL 37 \cite{chembl}, and Mcule (In Stock subset) \cite{mcule}.

We measure the total computation time, molecules per second (throughput), and total memory usage when using sparse fingerprints. To contextualize the gain from the latter, we also calculate how much memory the dense representation would use, and the memory savings. Results are summarized in Table~\ref{tab:large_scale_fingerprints}. \textit{scikit-fingerprints} is highly scalable to such libraries even on commodity hardware, both in terms of wall time and RAM requirements.

\begin{table}[]
\caption{Large-scale fingerprint computation.}
\resizebox{\textwidth}{!}{
\begin{tabular}{|c|c|c|c|c|c|c|}
\hline
\textbf{Dataset} & \textbf{\begin{tabular}[c]{@{}c@{}}Number of\\ molecules\end{tabular}} & \textbf{\begin{tabular}[c]{@{}c@{}}Total\\ time \end{tabular}} & \textbf{\begin{tabular}[c]{@{}c@{}}Molecules\\ per second\end{tabular}} & \textbf{\begin{tabular}[c]{@{}c@{}}Sparse matrix\\ memory\end{tabular}} & \textbf{\begin{tabular}[c]{@{}c@{}}Dense matrix\\ memory\end{tabular}} & \textbf{\begin{tabular}[c]{@{}c@{}}Memory usage\\ reduction ratio\end{tabular}} \\ \hline
COCONUT & $695$k & $15$ s & $46.3$k & $35$ MB & $1.3$ GB & $39 \times$ \\ \hline
ChEMBL & $2.9$M & $62$ s & $46.7$k & $145$ MB & $5.5$ GB & $39 \times$ \\ \hline
Mcule & $7.1$M & $1534$ s & $4.7$k & $312$ MB & $13.6$ GB & $45 \times$ \\ \hline
\end{tabular}
}
\label{tab:large_scale_fingerprints}
\end{table}

\section{Conclusion}

\textit{scikit-fingerprints} brings the breadth of chemoinformatics into the scikit-learn ecosystem. Building on RDKit, it unifies molecular fingerprints, filters, distance and similarity measures, applicability domain estimation, scaffold- and diversity-aware data splitting, fingerprint-aware hyperparameter tuning, virtual screening metrics, and curated benchmark datasets behind a single, consistent interface. Because every component is a scikit-learn transformer or a compatible function, an entire workflow - from a raw SMILES string to a serialized, deployable model - can be expressed as a composable pipeline and combined freely with the surrounding ecosystem. Alongside this uniformity, our focus on parallelization, mini-batching, and sparse representations keeps the library efficient, while its extensible design makes adding custom components straightforward.

\textit{scikit-fingerprints} is under active development, with new functionalities further expanding the depth and scope of the library. We are working towards integrating embeddings from pretrained neural networks, like CLAMP \cite{clamp} and CheMeleon \cite{chemeleon}, as molecular fingerprints based on representation learning \cite{neural_embeddings}. Other directions include adding clustering, such as Taylor-Butina \cite{butina}, and diversity picking like Ashton's algorithm \cite{maxmin}, as well as further train-test splitting strategies like DataSAIL \cite{datasail}. As a fully open-source project under the permissive MIT license, \textit{scikit-fingerprints} welcomes community contributions, and we hope it will continue to lower the barrier to reproducible, production-ready molecular machine learning.

\section{Data and Software Availability}

\begin{sloppypar}
All data used is publicly available and hosted on HuggingFace Hub at \url{https://huggingface.co/scikit-fingerprints/datasets}. \textit{scikit-fingerprints} is a free open source software under MIT license, available on GitHub: \url{https://github.com/MLCIL/scikit-fingerprints}.
\end{sloppypar}

\begin{acknowledgement}

The authors thank all contributors to the \textit{scikit-fingerprints} library, in particular Piotr Ludynia and Michał Stefanik. They also thank BIT Student Scientific Society for computational resources and organizational help.

\end{acknowledgement}





\bibliography{bibliography}

\end{document}